\documentclass[10pt,letterpaper]{article}

\usepackage{cogsci}
\usepackage{pslatex}
\usepackage{apacite}
\usepackage{graphicx}
\usepackage{algorithmic}
\usepackage{algorithm}
\usepackage{natbib}

\usepackage{subfigure}
\usepackage{colortbl}
\usepackage[rightcaption]{sidecap}

\title{Measuring Cultural Relativity of Emotional Valence\\and Arousal using Semantic Clustering and Twitter}
 
\author{{\large \bf Eugene Y. Bann (eugene@aeir.co.uk)} \\
 Advanced Emotion Intelligence Research (AEIR)
 \AND {\large \bf Joanna J. Bryson (J.J.Bryson@bath.ac.uk)} \\
 Department of Computer Science, University of Bath, United Kingdom}

\begin{document}

\maketitle

\begin{abstract}
Researchers since at least Darwin have debated whether and to what  extent emotions are universal or culture-dependent. However, previous studies have primarily focused on facial expressions and on a limited set of emotions. Given that emotions have a substantial impact on human lives, evidence for cultural emotional relativity might be derived by applying distributional semantics techniques to a text corpus of self-reported behaviour. Here, we explore this idea by measuring the valence and arousal of the twelve most popular emotion keywords expressed on the micro-blogging site Twitter. We do this in three geographical regions: Europe, Asia and North America. We demonstrate that in our sample, the valence and arousal levels of the same emotion keywords differ significantly with respect to these geographical regions --- Europeans are, or at least present themselves as more positive and aroused, North Americans are more negative and Asians appear to be more positive but less aroused when compared to global valence and arousal levels of the same emotion keywords. Our work is the first in kind to programatically map large text corpora to a dimensional model of affect.

\end{abstract}

\section{Introduction}

% DO WE NEED TO TALK ABOUT LINGUISTIC RELATIVITY (S-W HYP) (DON'T WANT
% TO GET INTO THAT CAN O' WORMS...)? -- leave that to the reviewers JJB

The question as to whether the experience and expression of emotions is universal or relative to specific cultures has resulted in a wide variety of studies, with theories ranging from the universality hypothesis to culture-specific facial expressions. Here we present evidence that culture is a necessary framework for researchers studying variation in emotions. Independent of the question of biological differences in the experience of emotions, it would be unsurprising if culture shapes our conscious perception, expression and experience of emotions, as has been hypothesised for other cognitive phenomena \citep{Hunt91,Fuhrman11}. Here, we use Latent Semantic Clustering on an emotional text corpus mined from Twitter to discern how the primary properties normally attributed to emotional keywords --- valence and arousal --- differ as the keywords are used in the same language (English) as exploited across different global regions.

\subsection{The Conceptualisation of Emotion Qualia}

{\em Emotion qualia} refers to the raw feel of an emotion. The actual phenomenon of a particular emotion experienced may differ according to each person's perception or understanding of that emotion, with perception being the result of the individual's past and hypothesised responses, unique to each human being. \citet{conceptual-emo} describes the act of conceptualising core affect, or in other words, why people attach emotion labels to the experience of emotion qualia. Since emotion keywords are constructed from conceptual knowledge about the world, emotions themselves may be concepts that humans begin learning in infancy and continuously extend and revise throughout life \citep{conceptual-act}. This repeated experience of labelling a combination of core affect and the context in which it occurs as an emotion provides training in how to recognise and respond to that emotion. In this sense, Barrett describes emotions as simulations. This skill of conceptualising core affect as an emotion could be a core aspect of emotional intelligence, in much the same way as conceptual thinking is core to cognitive intelligence. Each person learns the label in association with their unique experience, thus each person's conceptualisation of their emotional spectrum is unique. Cultures, formed of communicating individuals, may therefore also be unique if individual experiences vary somehow systematically. We base our analysis on this hypothesis. The reader should bear in mind that we are not analysing emotion keywords in particular, rather, we are analysing emotion \textit{conceptualisations}, or what cultures understand specific emotion keywords to mean, using Latent Semantic Clustering to infer these meanings.

\subsection{Core Affect}
Core affect is an emerging paradigm in affective neuroscience, and postulates a continuous approach to defining emotions \citep{circumplex}. Several core-affect, or \textit{circumplex} models have been proposed \citep[e.g.][]{ca1,ca2,ca3}, yet all have one thing in common: they represent emotions as a single point in a continuous space defined by two (or rarely three) dimensions. Different labels have been assigned to these two dominant dimensions by various theorists, such as {\em pleasure} and {\em engagement}, however most commonly, {\em valence} and {\em arousal} are chosen. Thus far, there has been no attempt to computationally pinpoint emotions or documents within a core affect model using `online' and `big' data; to date, research regarding the core affect model has either been theoretical \citep[e.g.][]{ca1}, or conducted via a limited survey \citep[e.g.][]{ca2}.

Core affect is one of two main theories regarding the representation of emotions, the other being the Basic Emotion model, however, neither has thus far received unequivocal support. Basic emotions could turn out to map to multiple subtypes of coherent emotion networks, but this implies we need to split basic emotion categories into further subtypes to better reflect these emotion networks \citep{hamann, bann12}. Here we extend this view and suggest that the core affect model enables us to quantify the properties of the basic emotions themselves. 

% It is our view that, in % keywords of linguistical approaches to emotion, the core affect model
% provides an increased level of granularity, and thus trumps the basic
% emotion model. -- JJB this is sufficiently controversial to leave
% to the discussion if at all.

\subsection{Previous Work}

% There has been several recent studies investigating the link between cultures and differing conceptualisations of emotions.

There is growing evidence that aspects of a person's psychology can be predicted from their language usage. In the 1990s, human semantics was shown to be recoverable from linguistic corpra independent of any further grounding \citep{lowe, BrysonMindSoc08}. Recent applications to individual psychology include discovering individual differences in personality \citep{personality}, discovering cultural change in moral beliefs \citep{bilovich}, as well as for emotion categorization \citep{cog11}. French discovered that co-occurrence techniques such as LSA does not detect personality from short text samples \citep{F2}, but do reveal that texts expressing particular emotions have a greater semantic similarity to corresponding exemplar words \citep{F1}.

A recent study by \citet{pnas} found significant evidence that facial expressions are indeed culture-dependent; that is, different cultures represent the same emotions differently. However, whether or not this is because they \textit{experience} different emotion qualia is another question. Using language, rather than facial expressions, as an accessor to emotion will enable a much more detailed and less ambiguous analysis, increasing significance by ``throwing more data at the problem" \citep[p.3]{more-data}.

Currently, there have been few attempts to analyse cultural differences using language semantics. Language plays a key role in how emotions are conceptualised (and thus perceived); Lindquist states ``language can be no more removed from emotion, than flour can be removed from an already baked cake" \citep[p.1]{powerful}. Recently, \citet{bannbryson12} demonstrated how conceptualisations of emotions can be inferred by performing Latent Semantic Analysis on a corpus of self-reported emotional tweets. Their DELSAR algorithm analysed 21,000 tweets each labelled with an emotion, and clustered each document in the corpus to its most similar corresponding emotion label using Latent Semantic Clustering. Here we use the same algorithm as the basis for our analysis.

\section{Corpus}

Typing emotion keywords into the Internet is increasingly becoming a significant technique for individual expression. There now exists a rich available source of information about emotions on the Internet, because so many people spend time expressing how they feel in blogs, forums, social networking websites and the like. We use data from the microblogging website Twitter to perform large-scale analysis of the language used in thousands of expressions of emotions within tweets. Acquiring a significantly larger corpus than \citet{bannbryson12}, we use the Gardenhose level of Twitter's streaming API\footnote{https://dev.twitter.com/docs/streaming-apis.} to create a corpus of 5,625,844 tweets\footnote{Having first removed 34,725 duplicate tweets. Corpus and code is available to download at www.aeir.co.uk/code.} collected between 19th October 2012 and 18th January 2013. Each emotion keyword (see selection criteria below) is given a five-minute streaming window in turn for the duration of the period, ensuring an even temporal distribution of Tweets is collected. Table~\ref{stats} describes our corpus, split by `cultural' region. We use the tweet's timezone as an indication of the corresponding user's geographical location; seeing as it is very unlikely that a Twitter user would select a timezone other than that which they reside in, it is somewhat safe to assume that this reflects the cultural origin of each user.

\begin{table}[!ht]
\begin{center} 
\caption[Distribution of tweets within our corpus]{Distribution of tweets within our corpus.}
\label{stats}
\vskip 0.12in
\begin{tabular}{lrrrr}
\hline
\textbf{Emotion} & \textbf{Asia} & \textbf{Europe} & \textbf{NA} & \textbf{All} \\
\hline
Angry & 12194 & 27070 & 61293 &200024 \\
Ashamed & 1008 & 5097 & 17107& 46486\\
Calm & 5975 & 10181 & 36681& 102827\\
Depressed & 3078 & 11615 & 43129&120473 \\
Excited & 30923 & 100792 & 292822&847679 \\
Happy & 149129 & 186709 & 730839& 2201874\\
Interested & 3527 & 9728 & 31891&86763 \\
Sad & 46351 & 83075 & 341912&966165 \\
Scared & 15435 & 42500& 194130 &517715 \\
Sleepy & 26031 & 10787 & 120473 &290666\\
Stressed & 2587 & 8774 & 41716& 109295 \\
Surprised & 3032 & 12454 & 56332 & 135877\\
\hline
\textbf{Total} & \textbf{299270} & \textbf{508782} & \textbf{1968325} & \textbf{5625844}\\
\hline
\end{tabular} 
\end{center} 
\end{table}

\noindent
\textbf{Region definitions.} We only include those timezones that have over 5000 tweets within our corpus. The Asia region consists of the timezones \textit{Kuala Lumpur, Beijing, Singapore, Jakarta, Bangkok, Hong Kong, Tokyo}; the Europe region consists of the timezones \textit{London, Amsterdam, Athens, Edinburgh, Dublin, Berlin, Paris}; the North American (NA) region consists of the timezones \textit{Eastern Time (US \& Canada), Central Time (US \& Canada), Mountain Time (US \& Canada), Pacific Time (US \& Canada)}. 

\noindent
\textbf{Selection of emotions.} As opposed to strictly using the basic emotions as identified by \citet{bannbryson12}, we use the most popular emotions that are used on Twitter, that is, those emotions that have the highest stream rate. Twelve emotions were selected that had a high rate and that equally divided into positive/negative and engaged/disengaged theoretical categories (see Table~\ref{theoretical}).

\noindent
\textbf{Subcorpus creation.} Each subcorpus is created using a limit of 1000 documents per emotion for all subcorpora to ensure consistency within our results; we chose 1000 as it is the lowest value in Table~\ref{stats}. To mitigate micro-temporal effects, if the number of documents for a particular emotion is significantly greater than 1000, we use a modulus function to extract 1000 documents equally spaced across the subcorpus --- for example, if a particular emotion in a particular subcorpus has 6000 documents, we take one document every six documents. We also create six control subcorpora so to compare our region-specific results with a baseline. We use the same modulus function to extract 1000 equally spaced tweets, but without any timezone clause, selecting six random starting points.

\section{Proposed Method}

We use DELSAR \citep[see Algorithm~\ref{DELSAR}]{bannbryson12} to generate the clustering matrix for each subcorpus --- the three regions Asia, Europe and NA, and six random controls.

\begin{algorithm}
%\small
\caption{\small DELSAR}
\label{DELSAR}
\begin{algorithmic}

\REQUIRE Corpus \textbf{C} and Keyword Set \textbf{K}, where each document in \textbf{C} is mapped to one emotion keyword, $emotion$, in \textbf{K} (through corpus generation)

\STATE Generate cosine document similarity matrix of LSC(\textbf{C}, \textbf{K})  ($document \times document$ similarity matrix)

\FOR{\textbf{each} \textit{emotion} $\in$ \textbf{K}} 

\FOR{\textbf{each} \textit{document} that has emotion \textit{emotion}} 
\STATE \textbf{delete} $emotion$ within the \textit{document}
\STATE Find the closest document $nearest$ where $nearest$ $\neq$ \textit{document}
\STATE Increment the count for the emotion that $nearest$ is labelled as in $emotion\_vector$
\ENDFOR \textbf{ each}
\RETURN $emotion\_vector$
\ENDFOR \textbf{ each}

\end{algorithmic}
\end{algorithm}

For each subcorpus, DELSAR uses LSA \citep{lsa} to create a document-document matrix of cosine similarities (\textit{Similarity Matrix}), in which similar documents are closer to one (i.e. the cosine of the angle between their vectors). It creates a clustering matrix that represents the corpus as an emotion-emotion matrix, describing how each emotion is similar to each other emotion.

% , or in other words,
% the amount each emotion is composed of each other emotion. For
% example, the emotion \textit{joyful} is composed of, say, 30\% of
% \textit{joyful} documents, 10\% of \textit{excited} documents, and 6\%
% of documents for each of the other 10 emotions. -- JJB I'm not clear
% on how one emotion can be composed of another -- you mean that
% tweets contain both keywords? Anyway, let's leave it out, they will be
% used to LSA & you'll need the space.

All analysis was performed on a 64-bit Intel Core i5 CPU 2x2.67GHz with 4GB RAM using the \textsc{gensim} framework for Python \citep{radim} to create LSA spaces. For all tasks, we use a dimension of 36 and use Log-Entropy normalisation as our Association Function, found to generate optimal results \citep{lsa-weight} and recommended for LSA \citep{lsa}.

\subsection{Valence and Arousal}

Here we take valance to mean the theoretical positive or negative attribution of an emotion keyword, and similarly arousal to mean the implied level of engagement. We should use the keywords \textit{theoretical} valence and \textit{theoretical} arousal as we are measuring emotion keywords relative to their generally accepted categorisation, although there does seem to be consistency in these categorisations between theorists. Table~\ref{theoretical} shows the theoretical definitions of our keywords, accumulated using several circumplex models of affect \citep{ca1,ca2,ca3}.

\begin{table}[!ht]
\begin{center} 
\caption[Valence and arousal categorisation of the twelve emotion keywords analysed]{Valence and arousal categorisation of the twelve emotion keywords analysed.}
\label{theoretical}
\vskip 0.12in
\begin{tabular}{l*{3}{p{2.2cm}}r}
\hline
\textbf{Emotion}  & \textbf{Valence} & \textbf{Arousal} \\
\hline
Angry & Negative & Engaged  \\
Ashamed & Negative & Disengaged  \\
Calm  & Positive & Disengaged \\
Depressed & Negative & Disengaged \\
Excited  & Positive & Engaged \\
Happy  & Positive & Disengaged \\
Interested & Positive & Engaged \\
Sad  & Negative & Disengaged \\
Scared & Negative & Engaged \\
Sleepy  & Positive & Disengaged \\
Stressed & Negative & Engaged \\
Surprised  & Positive & Engaged \\
\hline
\end{tabular} 
\end{center} 
\end{table}

We calculate the valence and arousal levels of each emotion for each subcorpus as follows. First, we run DELSAR on the subcorpus to generate clustering vectors for each emotion. Each emotion's valence is then calculated as the number of \textit{positive} elements within its vector, as defined in Table~\ref{theoretical}, divided by the total number of documents across all elements (which will always be 1000), or in other words, the percentage of \textit{positive} elements within its vector. Similarly, each emotion's arousal is calculated as the percentage of \textit{engaged} elements within its vector, again as defined in Table~\ref{theoretical}. We then normalise each valence and arousal value by taking away the average valence and arousal value, respectively, for \textit{all} subcorpora analysed --- Asia, Europe and NA regions and the six control subcorpora. This ensures relativity of the resulting circumplex model between these analysed groups; these groups can now be compared to one another to establish similarities and differences between them.

\section{Results}

\begin{figure*}[ht]
\begin{center}
\includegraphics[width=166mm]{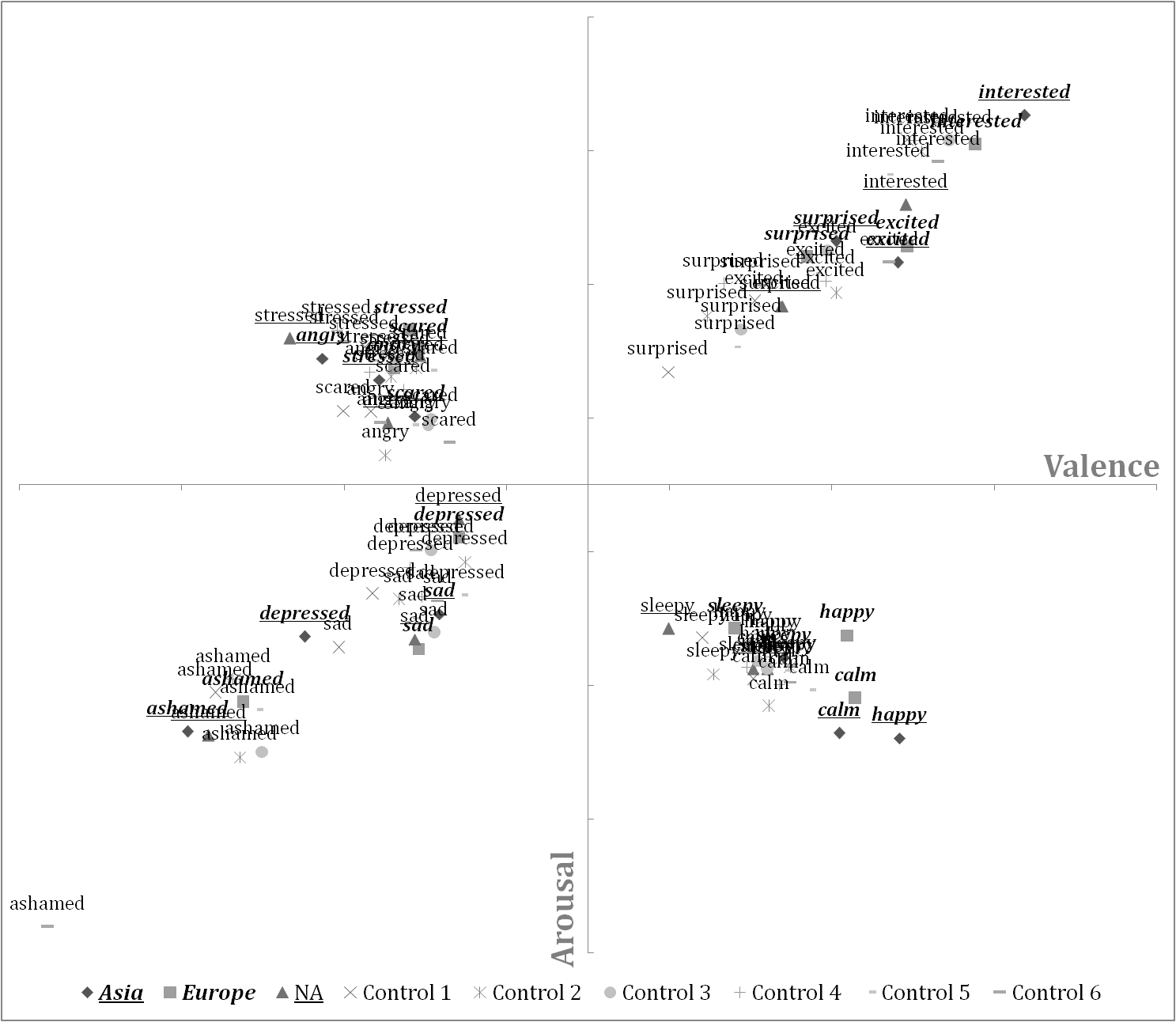}
\end{center}
\caption[Circumplex of three regions and six controls]{Circumplex of three regions and six controls.}
\label{plot}
\end{figure*}

\begin{figure}[ht]
\begin{center}
\includegraphics[height=75mm]{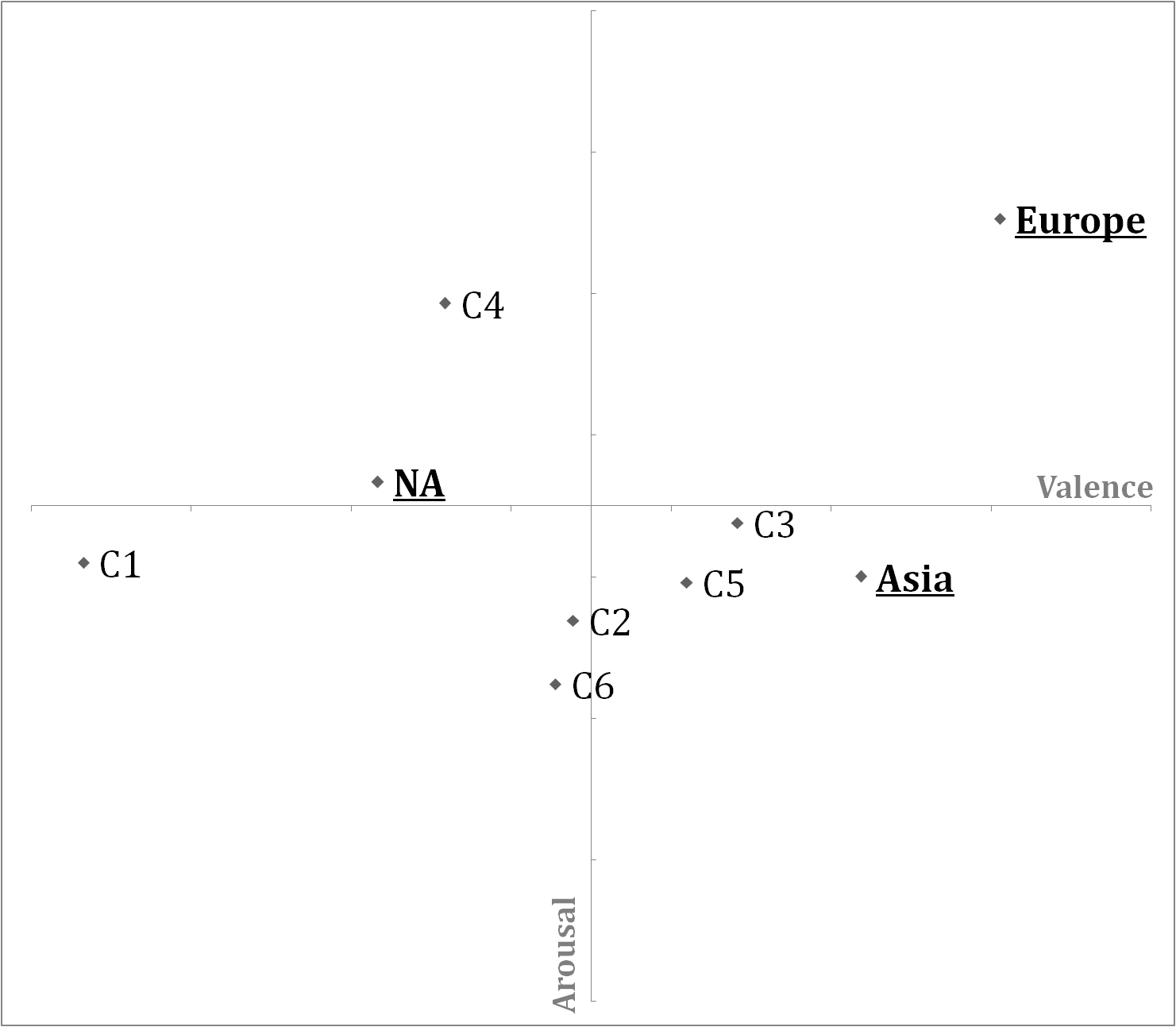}
\end{center}
\caption[Aggregate theoretical positivity and engagement for each subcorpus]{Aggregate theoretical positivity and engagement for each subcorpus.}
\label{plot2}
\end{figure}

Figure~\ref{plot} shows a plot of our circumplex of selected subcorpora. We can see that some emotions are more tightly packed than others, and interestingly, that low-valence-high-arousal and high-valence-low-arousal emotions are much more universally similar when compared to the other two quadrants of the circumplex. In order to visualise each separate region more clearly we illustrate the aggregate theoretical positivity and engagement for each subcorpus, shown in Figure~\ref{plot2}. This clearly illustrates that our three regions do indeed have different conceptualisations of the same emotion keyword; we see that the region \textit{Europe} is a much more \textit{positive} and \textit{engaged} culture; in other words, Europeans find the same emotion keywords to be more positive and engaging when compared to other cultures and indeed our control samples. Also, we discover that Asians find the same emotion keywords to be somewhat more positive, and North Americans somewhat more negative, with negligible arousal differences.

\begin{figure}[ht]
\begin{center}
\includegraphics[height=75mm]{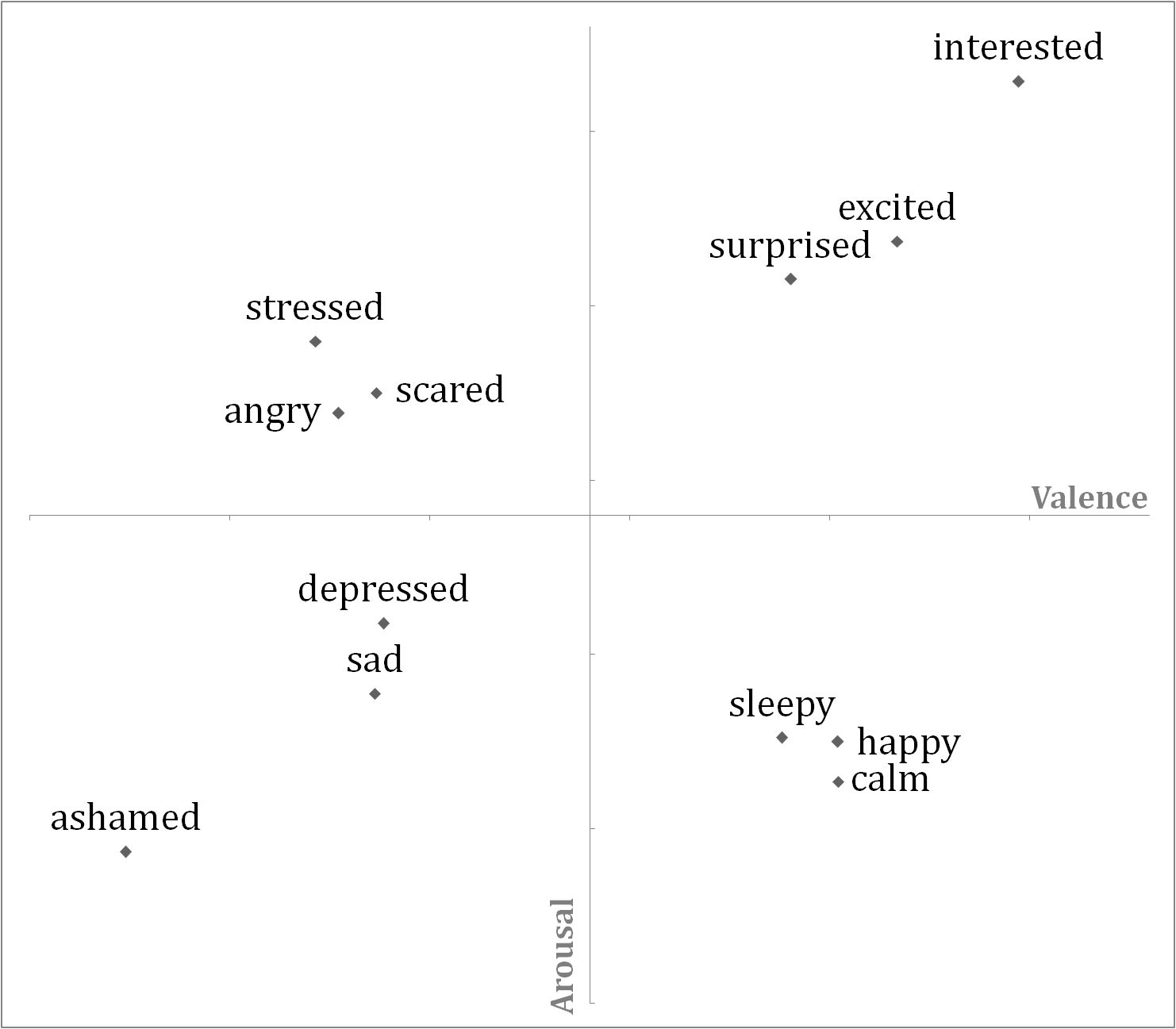}
\end{center}
\caption{Centroid emotion circumplex.} 
\label{plot3}
\end{figure}

In order to analyse how tightly packed our emotion clusters are in Figure~\ref{plot}, we conducted K-Means cluster analysis to determine the centroids for each emotion, calculating the distances of each emotion to its centroid. We plot our centroids, shown in Figure~\ref{plot3}, resulting in a circumplex that could be thought of as a universal emotion circumplex, illustrating what people think emotions to be, relative to each other emotion. We can see that the emotions \textit{scared}, \textit{depressed} and \textit{sad} have a very similar valence, yet varying arousal levels; so too do the emotions \textit{sleepy} and \textit{sad}. We can also see, albeit less definitively, that the emotions \textit{stressed} and \textit{surprised} have a similar arousal level, but opposite valence; so too do the emotions \textit{sad} and \textit{sleepy}.

In order to identify which emotions have the most and least similar conceptualisations across cultures, we calculate the distance of each emotion to its respective centroid for each region, and calculate the sum of these distances for each emotion across all subcorpora, shown in Table~\ref{sdev}. We discover that the emotions \textit{sad} and \textit{stressed} have the most similar conceptualisations across all cultures; in other words, people understand these two emotions to mean the same thing independent of culture. Similarly, we find that the emotions \textit{surprised} and \textit{depressed} have the most widely varying conceptualisations across cultures; in other words, different cultures have very different valence and arousal attributions towards these two emotions. Note that we do not include the emotion \textit{ashamed} in the top two due to a strange anomaly in control group 6 which skews an otherwise relatively tight cluster.

\begin{table}[!ht]
\begin{center} 
\caption[Sum of subcorpus distances to respective centroids]{Sum of subcorpus distances to respective centroids.}
\label{sdev}
\vskip 0.12in
\begin{tabular}{lr|lr}
\hline
\textbf{Emotion} & \textbf{Distance} & \textbf{Emotion} & \textbf{Distance} \\
\hline

Sad & \textbf{17.94} &Scared & 23.27\\
Stressed & \textbf{19.66}&Happy & 26.10\\
Calm & 20.86&Excited & 27.00\\
Interested & 22.72&Depressed & \textbf{29.56}\\
Angry & 23.18&Surprised & \textbf{32.89}\\
Sleepy & 23.19&Ashamed & \textit{40.70}\\
\hline
\end{tabular} 
\end{center} 
\end{table}

\subsection{Discussion}

We would expect that the control groups would be tightly clustered around the centre of the circumplex in Figure~\ref{plot2}, and for the most part, they are. The exceptions are control groups one and four, possibly due to the fact the the corpus is skewed in favour of tweets originating from NA (see Table~\ref{stats}); this is somewhat verified by their closeness to the NA subcorpus. Other than these anomalous subcorpora, the circumplex does illustrate how different cultures significantly conceptualise emotions differently, in keywords of valence and arousal. Interestingly, there are certain emotions in certain regions that stick out of our analysis. One example is the emotion \textit{depression}; Asians find this emotion much more negative than all other cultures and control groups. This could be due to cultural differences such as coping strategies \citep{asians}. Another example concerns the emotions \textit{happy} and \textit{calm}; Europeans and Asians find these emotions much more positive than North Americans and all control groups. Another suggests that Asians find \textit{interest} a very positive and aroused emotion, compared to North Americans who conceptualise the same emotions, relatively, as negative and disengaged.

% I think we need to work to make controls that are drawn evenly from
% all time zones, or at least from our global regions! let's submit
% it like this, but you should maybe do the new controls?

\subsection{Limitations}

We document several limitations of our approach. Firstly, our database may still contain duplicate tweets, as some users duplicate tweets by appending, for example, a number at the end, making them unique from one another. Second, our modulus function does not take an even sample for our control groups at the country level, so they may be skewed in
favour of countries with a higher frequency of documents within the database (our corpus on the whole is in fact skewed in favour of NA). Thirdly, we assume that the emotion keywords we have selected are in fact emotion \textit{qualia} as opposed to adjectives. Fourth, our corpus is essentially a snapshot in time and may reflect, for example, the political or economic climate at the time, or skew due to global events such as the US election. Finally, our corpus consists entirely of English tweets, which skews our results in favour of Western cultures; our Asia, and to some extent, Europe subcorpora may not be entirely representative of their respective cultures as we disregard all native languages other than English. In addition, the subpopulations of those regions who choose to use Twitter, and do so in English, may be a biased sample.

\section{Conclusions}

Emotions are being increasingly expressed online, and being able to understand these emotions is rapidly becoming a concern of AI and Cognitive Science. By mapping culture-specific emotion circumplexes, we hope to be better able to understand culture-specific perceptions or even experience of emotions. From the work presented here we can
conclude the following:\\
\textbf{Emotional semantics depends on culture.} The same emotion keyword in one culture may describe different valence and arousal properties in another. This seems to be more true of some keywords than others, and could be critical where, for example, a significantly differing conceptualisation of the emotion \textit{depression} would require a different understanding and response.\\
\textbf{Emotions vary by geographic region.} Europeans are more likely to express positiveness and engagement. Asians are also more positive than North Americans, both relative to each other and to the control subcorpora. Note that this may reflect cultural differences in the public expression of emotion rather than its actual qualia --- our method cannot disambiguate these.\\ 
\textbf{Some emotions do seem to be conceptualised universally.} The emotion keywords \textit{sad} and \textit{stressed} have the same conceptualisation across cultures, whereas cultures have the most disagreement regarding the conceptualisation of \textit{surprised}.
\\\\
\noindent
We hope that our research paves the way for a better understanding of how language can be used to identify specific properties of emotions, and we encourage the reader to verify our results by downloading our code and corpus at http://www.aeir.co.uk/code.

% \section{Future Work}

% Using our results and methodology, we will be building a web page classifier to be able to ``emotionally read" text, 
%relative to cultural differences of perception of the same article. Further work could focus on splitting up the largest of the 
%cultural regions selected, such as the differences between the east and west coasts of the United States. We are also 
%researching how time affects emotion conceptualisations using temporal, rather than cultural, subcorpora.

\bibliographystyle{apacite}

\setlength{\bibleftmargin}{.125in}
\setlength{\bibindent}{-\bibleftmargin}

\bibliography{CogSci_Template}

\end{document}